\documentclass[sigconf]{acmart}

\usepackage{subfig}

\AtBeginDocument{%
  }

\acmSubmissionID{9465}

\usepackage{graphicx}
\usepackage{makecell}

\begin{document}

\title{Interactive Interface For Semantic Segmentation Dataset Synthesis}

\author{Ngoc-Do Tran}
\authornote{Both authors contributed equally to this research.}
\affiliation{%
  \institution{University of Science}
  \city{Ho Chi Minh}
  \country{Vietnam}
}
\affiliation{%
  \institution{Vietnam National University}
  \city{Ho Chi Minh}
  \country{Vietnam}
}
\email{}
\author{Minh-Tuan Huynh}
\authornotemark[1]
\affiliation{%
  \institution{University of Science}
  \city{Ho Chi Minh}
  \country{Vietnam}
}
\affiliation{%
  \institution{Vietnam National University}
  \city{Ho Chi Minh}
  \country{Vietnam}
}
\email{}
\author{Tam V. Nguyen}
\orcid{0000-0003-0236-7992}
\affiliation{%
  \institution{University of Dayton}
  \city{Dayton}
  \state{Ohio}
  \country{US}
}
\email{}
\author{Minh-Triet Tran}
\orcid{0000-0003-3046-3041}
\affiliation{%
  \institution{University of Science}
  \city{Ho Chi Minh}
  \country{Vietnam}
}
\affiliation{%
  \institution{Vietnam National University}
  \city{Ho Chi Minh}
  \country{Vietnam}
}
\email{}
\author{Trung-Nghia Le}
\orcid{0000-0002-7363-2610}
\affiliation{%
  \institution{University of Science}
  \city{Ho Chi Minh}
  \country{Vietnam}
}
\affiliation{%
  \institution{Vietnam National University}
  \city{Ho Chi Minh}
  \country{Vietnam}
}
\email{}
\authornote{Corresponding author. Email: ltnghia@fit.hcmus.edu.vn}




\renewcommand{\shortauthors}{?}

\begin{teaserfigure}
    \centering
    \includegraphics[width=\textwidth]{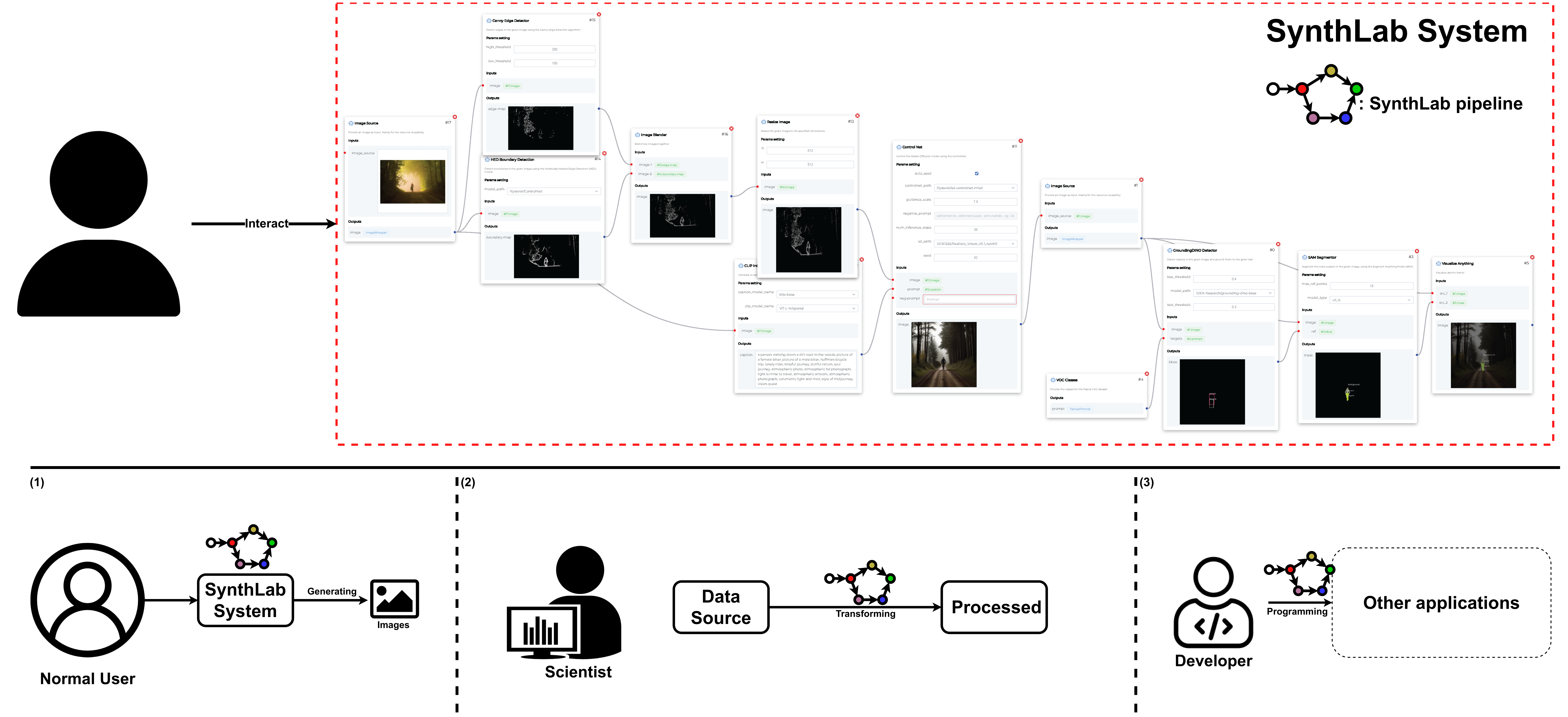}
    \caption{Through an interactive interface, SynthLab enables end-users to create their data generation pipeline quickly via drag-and-drop actions. SynthLab's motivation is to be a user-friendly environment for everyone and a powerful platform that can provide solutions for different groups of users, from normal users (1), and scientists (2) to developers (3). SynthLab includes generative modules that support generating on-demand high-quality images. On the other hand, modules in SynthLab can be connected and execute many different AI tasks from different areas. With the modular design, SynthLab's pipeline is not only usable in our system but also executable in the developing environment helping developers faster build up their app and go further with other groups of end-users.}
    \label{fig:system-overview}
\end{teaserfigure}

\begin{abstract}
  The rapid advancement of AI and computer vision has significantly increased the demand for high-quality annotated datasets, particularly for semantic segmentation. However, creating such datasets is resource-intensive, requiring substantial time, labor, and financial investment, and often raises privacy concerns due to the use of real-world data. To mitigate these challenges, we present SynthLab, consisting of a modular platform for visual data synthesis and a user-friendly interface. The modular architecture of SynthLab enables easy maintenance, scalability with centralized updates, and seamless integration of new features. Each module handles distinct aspects of computer vision tasks, enhancing flexibility and adaptability. Meanwhile, its interactive, user-friendly interface allows users to quickly customize their data pipelines through drag-and-drop actions. Extensive user studies involving a diverse range of users across different ages, professions, and expertise levels, have demonstrated flexible usage, and high accessibility of SynthLab, enabling users without deep technical expertise to harness AI for real-world applications.  
  
\end{abstract}



\maketitle

\section{Introduction}

Recent progress in generative artificial intelligence (AI), particularly in the domain of image generation, has demonstrated substantial potential to revolutionize various sectors~\cite{ho2020denoising, song2021denoising, nichol2021improved, nichol2021glide, sauer2023stylegan, kang2023scaling, rombach2022highresolution, podell2024sdxl, sd3}. These technologies enable the synthesis of novel and creative content, effectively translating conceptual ideas into tangible outputs. Two notable developments have emerged as pivotal in this landscape: ChatGPT~\cite{an2023chatgpt}, recognized as a leading model for text generation, and Stable Diffusion~\cite{rombach2022highresolution}, distinguished as the premier open-source text-to-image model. Together, these innovations are expanding the scope of digital interaction and creative expression, enhancing the accessibility and adaptability of AI for a global user base. Furthermore, text-to-image models derived from Stable Diffusion~\cite{nichol2021glide, rombach2022highresolution, podell2024sdxl, sd3} have garnered significant interest within the research and user communities, increasing their visibility and applicability among end users.

Despite these advancements, individuals with limited or no advanced technical expertise often encounter challenges in utilizing these technologies due to their inherent complexity. Consequently, there is a pressing need to develop simplified interfaces that integrate these methods, thereby broadening the accessibility of AI models to a wider audience. Such an interface would serve as a conduit between sophisticated AI systems and non-expert users, enabling seamless incorporation of AI into diverse workflows irrespective of technical proficiency. By providing intuitive design and user-centric functionalities, this tool would empower both individuals and organizations to harness the full potential of generative AI technologies, fostering innovation and improving operational efficiency across multiple domains.

In this paper, we introduce SynthLab, a novel platform designed for visual data synthesis and annotation, featuring a modular platform and an accessible user interface. The backend of SynthLab comprises a modular architecture engineered to encapsulate algorithms and AI models, standardizing their implementation. Each module adheres to a uniform interface, facilitating interoperability and enabling the construction of a cohesive data pipeline. These modules address distinct computer vision functions, such as image generation, segmentation, and object detection, thereby enhancing the system’s flexibility and adaptability. This modular design supports straightforward maintenance, scalability through centralized updates, and the seamless incorporation of new functionalities.

\begin{figure*}[t!]
    \centering
    \includegraphics[width=\textwidth]{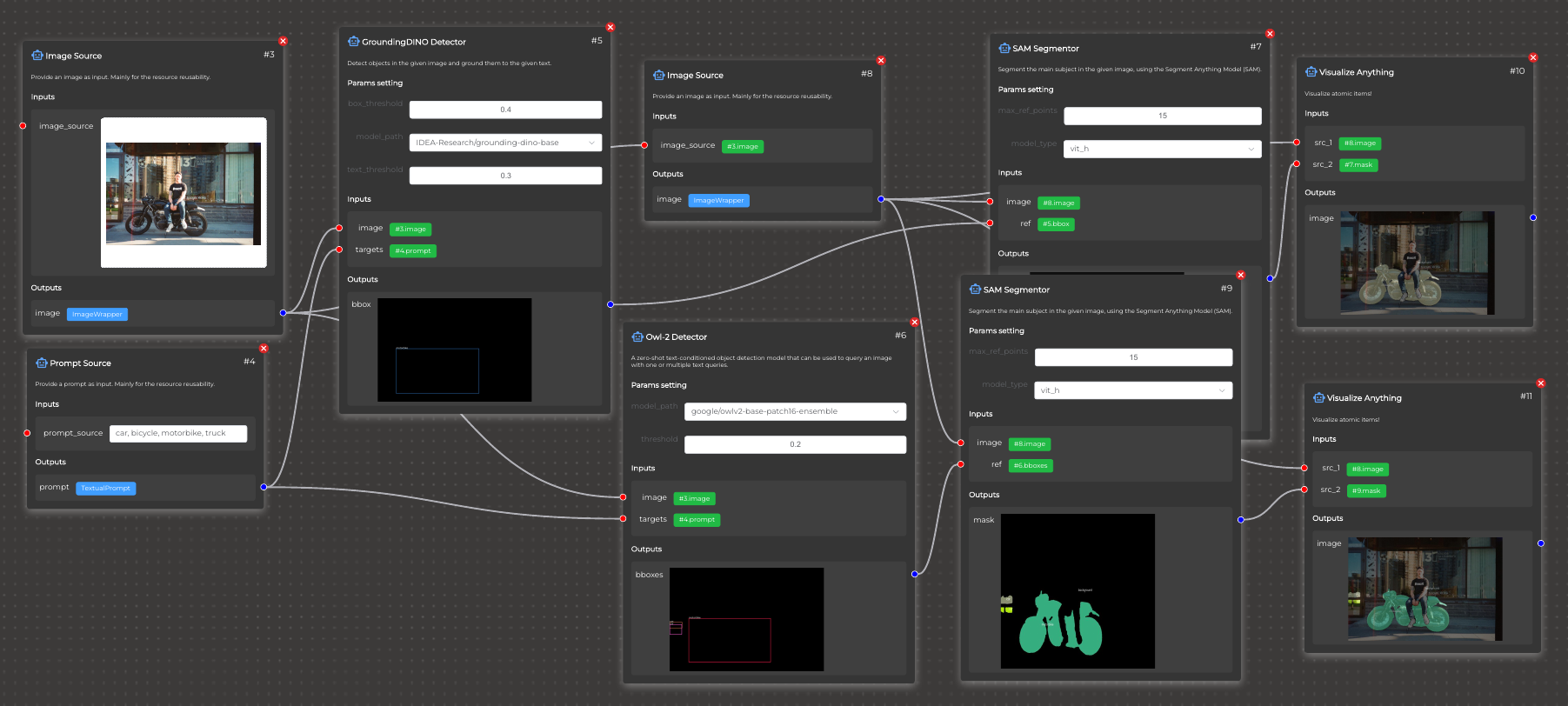}
    \caption{This pipeline was created by an undergraduate student in Task 2. The pipeline, involving GroundedSAM \cite{ren2024grounded} based method using two different zero-shot detection methods (e.g., GroundingDINO \cite{liu2023grounding} and OWLv2 \cite{minderer2024scaling}), followed by Segment Anything Model (SAM) \cite{kirillov2023segany}, achieves the highest confident output mask for the final dataset.}
    \label{fig:pipeline-khai}
\end{figure*}

Complementing the backend, SynthLab’s interactive front-end interface delivers an intuitive, hands-on experience, allowing users to engage with the platform’s core features without requiring advanced technical knowledge. The system empowers users to rapidly assemble data pipelines using a drag-and-drop mechanism, fine-tune module hyperparameters, and execute workflows with sample datasets (see Figure~\ref{fig:pipeline-khai}). Notably, user-defined pipelines can be deployed within a client-server environment, exported for private execution on user-specific data, or adapted for alternative applications. This versatility enhances accessibility, making SynthLab appealing to a diverse audience seeking to leverage AI technologies without the prerequisite of developing such systems from the ground up.

To evaluate SynthLab’s efficacy, we conducted comprehensive user studies across various tasks. Experimental outcomes underscore the platform’s novelty and utility in optimizing data workflow performance. SynthLab not only demonstrates exceptional ease of use but also delivers a high-quality, interactive user experience. By integrating advanced synthetic image generation models accessible via a web interface, SynthLab showcases robust data synthesis capabilities, enabling a broad spectrum of users to effectively apply the platform to their computer vision objectives.

In summary, our contributions are as follows:

\begin{itemize}
    \item We introduce SynthLab, a novel data processing and generating platform, which includes multiple modules from different areas, to address some real-world data issues systematically.

    \item SynthLab consists of a modular platform where each module handles distinct functionality but share the same interface, allowing the connectivity between modules, and forming a data pipeline. This allows to easily maintain and integrate new features. 

    \item SynthLab helps end-users create their data processing workflows quickly without requiring AI knowledge and programming skills via drag-and-drop actions in a friendly interactive interface.

    \item We discuss potential applications of our system through extensive experiments and user studies.
\end{itemize}

\begin{figure*}[t!]
    \centering
    \includegraphics[width=\linewidth]{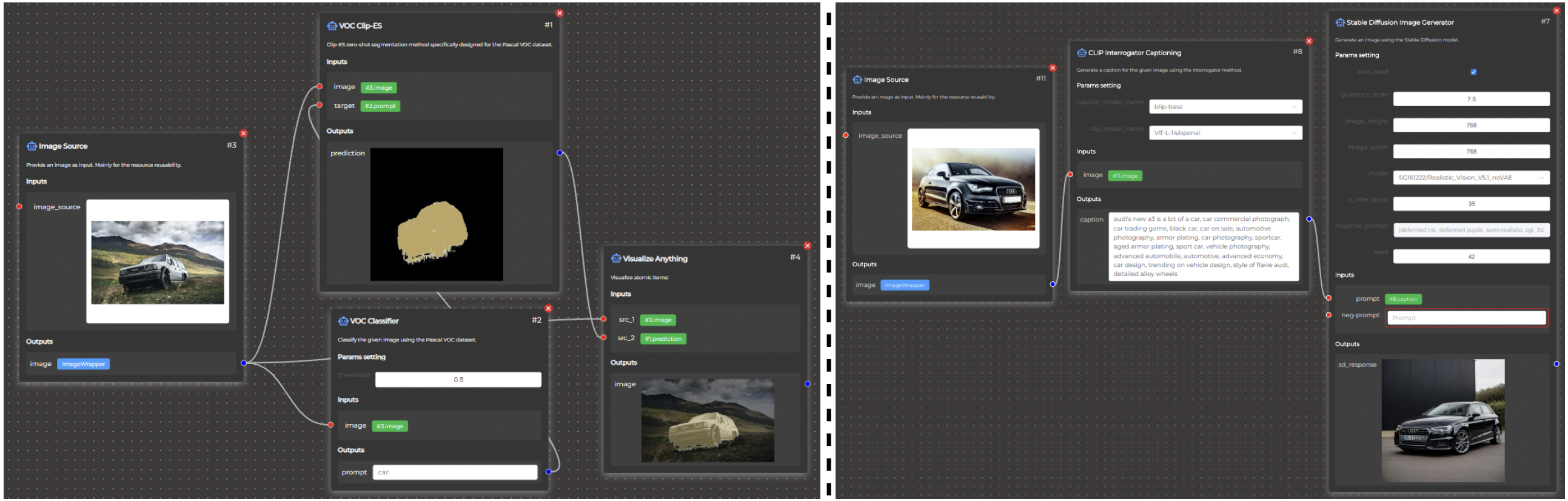}
    \caption{SynthLab pipeline can execute single computer vision tasks like classification, segmentation, and image generation. Additionally, with high compatibility, these modules in SynthLab can be connected together to create a more complex workflow.}
    \label{fig:pipeline-customization}
\end{figure*}

\section{Related Work}

Recent advancements have led to the development of accessible platforms designed to streamline the creation of machine learning workflows, minimizing the need for extensive programming knowledge. Platforms such as Google’s TensorFlow Playground\footnote{\url{https://playground.tensorflow.org}} and IBM’s Watson Studio\footnote{\url{https://www.ibm.com/products/watson-studio}} offer intuitive interfaces that enable users, including those without specialized expertise, to construct and visualize machine learning models. However, these tools are typically constrained to predefined workflows or specific model architectures, despite their robust visualization capabilities.

Visual programming environments \cite{burnett1995visual}, such as Node-RED \footnote{\url{https://nodered.org/}} and Orange \footnote{\url{https://orangedatamining.com/}}, leverage drag-and-drop interfaces to facilitate the construction of workflows through interconnected nodes. These nodes represent data inputs, processing operations, and outputs, providing an intuitive platform for applications in IoT, data science, and machine learning. This node-based paradigm supports the creation of complex workflows without requiring code, enhancing accessibility and clarity.

In the domain of computer vision and image generation, tools like NVIDIA's DIGITS\footnote{\url{https://developer.nvidia.com/digits}} and Runway ML\footnote{\url{https://runwayml.com/}} provide user-friendly interfaces with prebuilt models, often based on architectures like Stable Diffusion or deep learning networks. These platforms enable users to refine and deploy models for creative tasks without necessitating a detailed understanding of the underlying systems. Notably, ComfyUI\footnote{\url{https://github.com/comfyanonymous/ComfyUI}}, a node-based interface for Stable Diffusion \cite{saharia2022photorealistic}, exemplifies this trend. Adopted by StabilityAI for internal testing of Stable Diffusion features, ComfyUI’s flexibility in node manipulation underscores its utility, further evidenced by StabilityAI’s recruitment of its developers for internal tool development.

Similarly, AUTOMATIC1111\footnote{\url{https://github.com/AUTOMATIC1111/stable-diffusion-webui}}, an open-source web interface for Stable Diffusion \cite{saharia2022photorealistic}, prioritizes ease of use with a tabbed layout supporting image generation, inpainting, and upscaling. While it offers a robust feature set, including preconfigured workflows and prompt editing, its customization options are less extensive than ComfyUI’s. InvokeAI\footnote{\url{https://github.com/invoke-ai/InvokeAI}}, another node-based platform, provides a streamlined interface with enhanced functionalities, such as prompt engineering, model management, and image inpainting/outpainting, alongside the ability to integrate multiple models for improved outcomes, making it particularly suitable for users accustomed to visual workflow systems.

\section{Proposed System}

\subsection{Fundamental Design}

SynthLab is the back-side platform with a simple yet effective design. The platform acts as a core logic and provides artifacts for the interface. Furthermore, SynthLab is also designed to be used from the end-client-side where the defined workflow from the interface can be used on their own hardware, and process their data privately, and systematically.

\begin{figure}[t!]
    \centering
    \includegraphics[width=\linewidth]{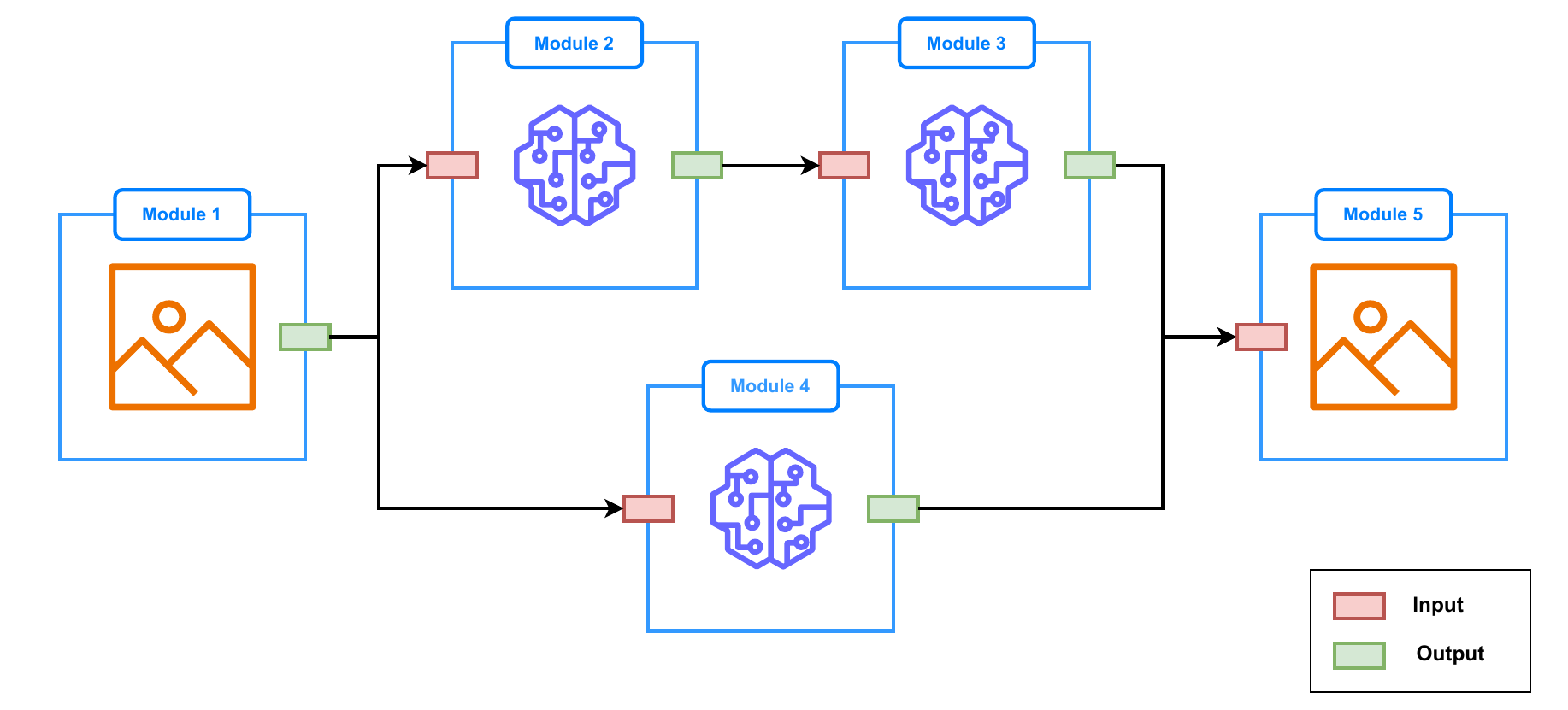}
    \caption{Fundamental design of SynthLab.}
    \label{fig:synthlab-node-pipeline}
\end{figure}

SynthLab standardizes AI models and algorithms into modules that execute by receiving input, executing, and returning output. As a high-level framework, the input and output of a module are user-readable, simply text, image, or mask (for other computer vision tasks). That led to the idea that we can use the output of one module to be the input of the others if they are the same data type (Figure \ref{fig:synthlab-node-pipeline}). 

Based on that basic idea, we initially implemented more than 50 SynthLab modules of different areas from vision generative, extractive models, language models, etc, and all of them work systematically together. Some module tasks that were added to SynthLab are Image Generation, text-to-image (T2I), image-to-image (I2I), and fundamental computer vision extractive tasks like image captioning, classification, and object detection to segmentation.

SynthLab empowers users with the ability to combine multiple modules within a single pipeline (Figure \ref{fig:pipeline-customization}). This modular approach allows for the integration of various state-of-the-art techniques to enhance the quality and diversity of synthetic data. This combination capability not only broadens the range of synthetic data that can be generated but also enables the creation of highly customized datasets that meet specific research or application needs.

\begin{figure*}[t!]
    \centering 
    \includegraphics[width=\linewidth]{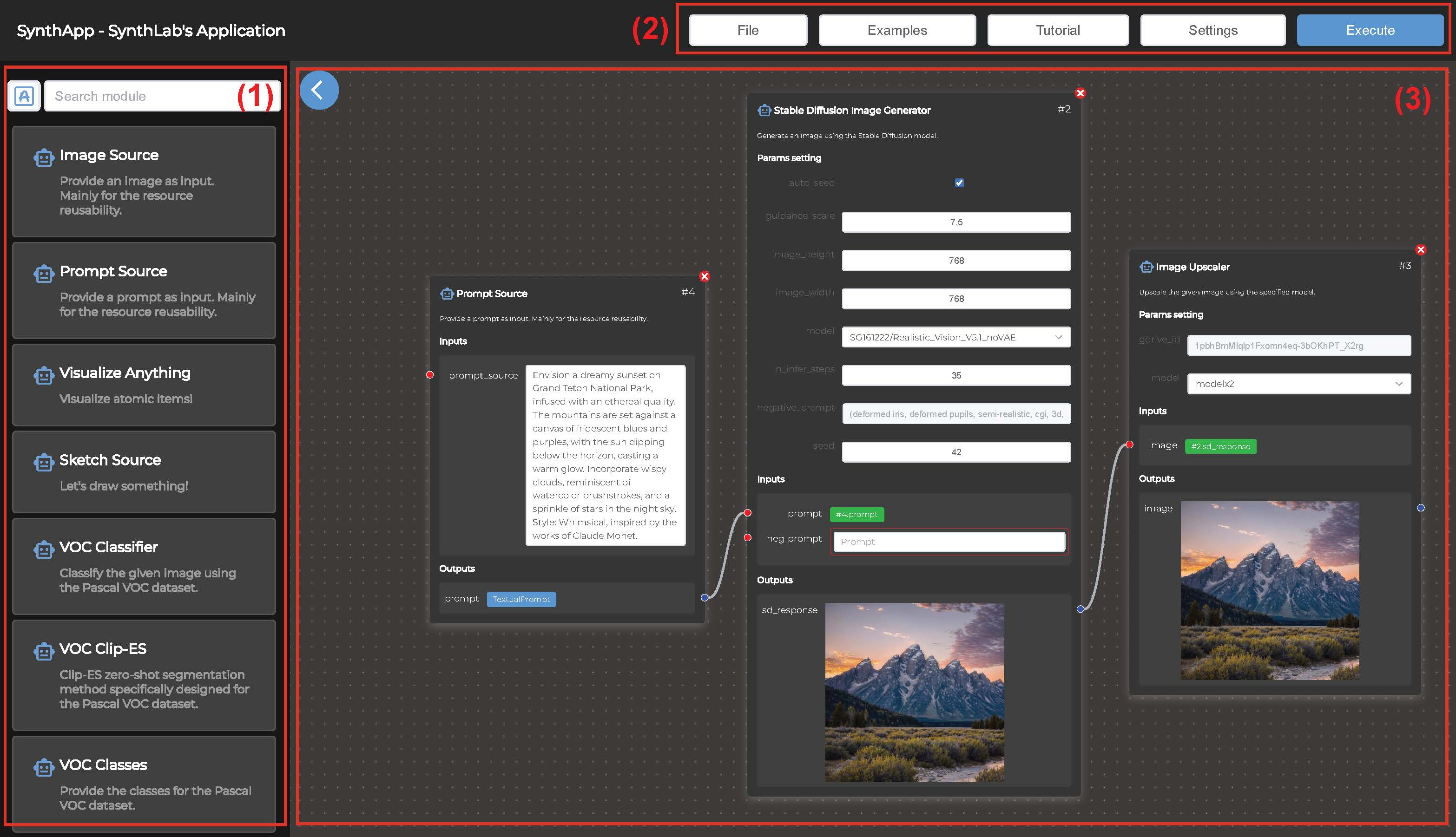}
    \caption{SynthLab's interface includes three main parts, the modules pool area (1), the settings area (2), and the playground (3).}
    \label{fig:synthapp-interface}
\end{figure*}

\begin{figure}[t!]
    \centering
    \includegraphics[width=\linewidth, trim={0 5cm 0 4cm},clip]{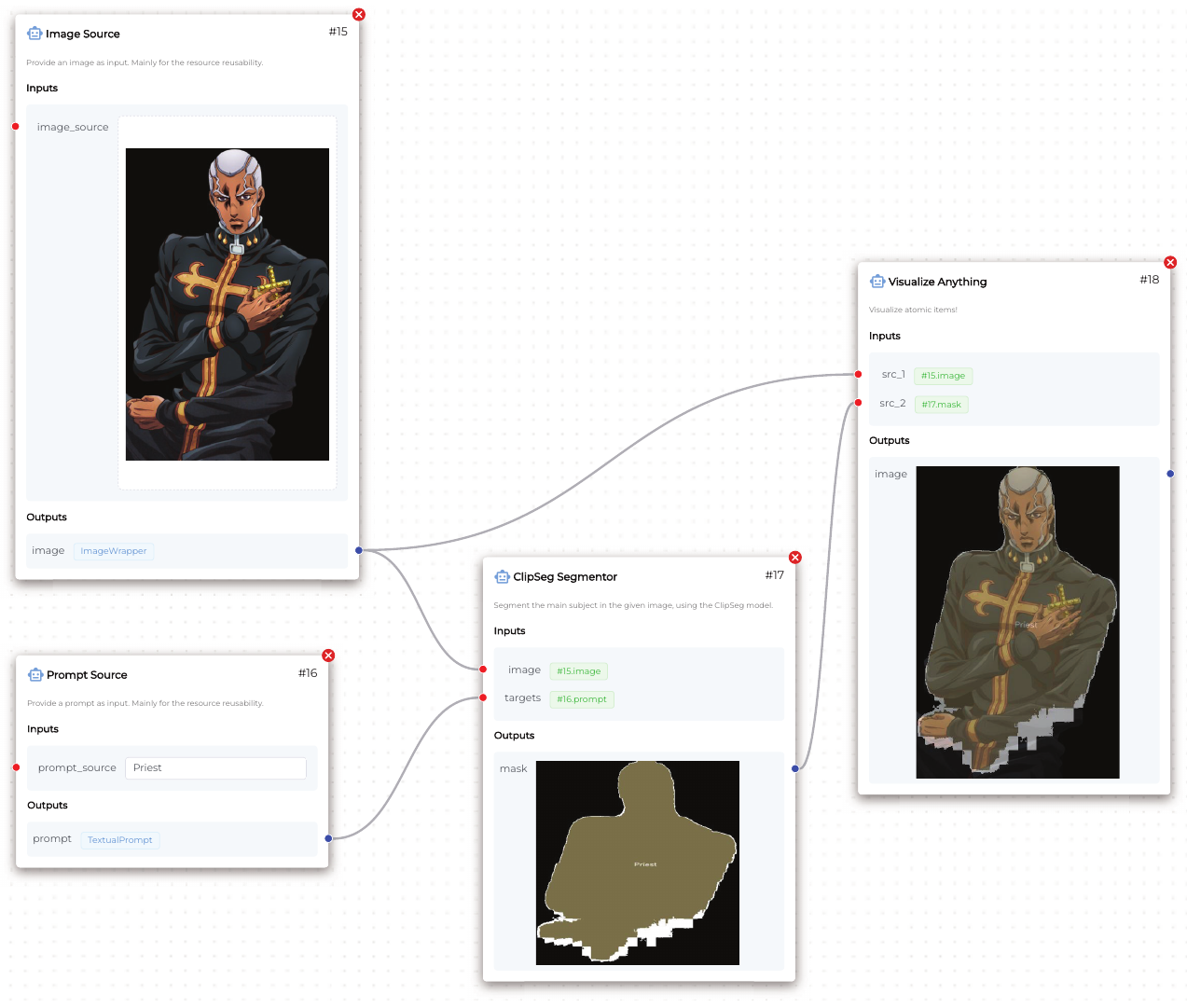}
    \caption{Pipeline customization is flexible in SynthLab.}
    \label{fig:pipeline-customization}
\end{figure}

\subsection{Node-based User Interface}

SynthLab's UI is created based on ComfyUI, our web application is a node-based interface. Node-based interfaces are often found in tools specializing in 3D design and effects such as Maya and Blender. In this type of interface, each node stands for a module in SynthLab. A node whose value is passed from the input points and returns the processed value to the output point.

The SynthLab's interface includes three main parts, the modules pool area (1), the settings area (2), and the playground (3) (Figure \ref{fig:synthapp-interface}. The module area shows all the modules that are supported by SynthLab, each includes a short description and a search bar for searching or quickly filtering the modules by labels (tasks related to a module). The settings area includes some options where the user can configure some options the the app, disable dark mode, or modify the weight of the edge, connected between nodes. The playground area is an open area where all the workflow of users is shown. To use the app, users just simply need to drag the module from the module area and drop it into the playground, then, the user can interact, submit sample input, and fine-tune the hyperparams of each module created.

\begin{figure*}[t!]
    \centering
    \includegraphics[width=\textwidth]{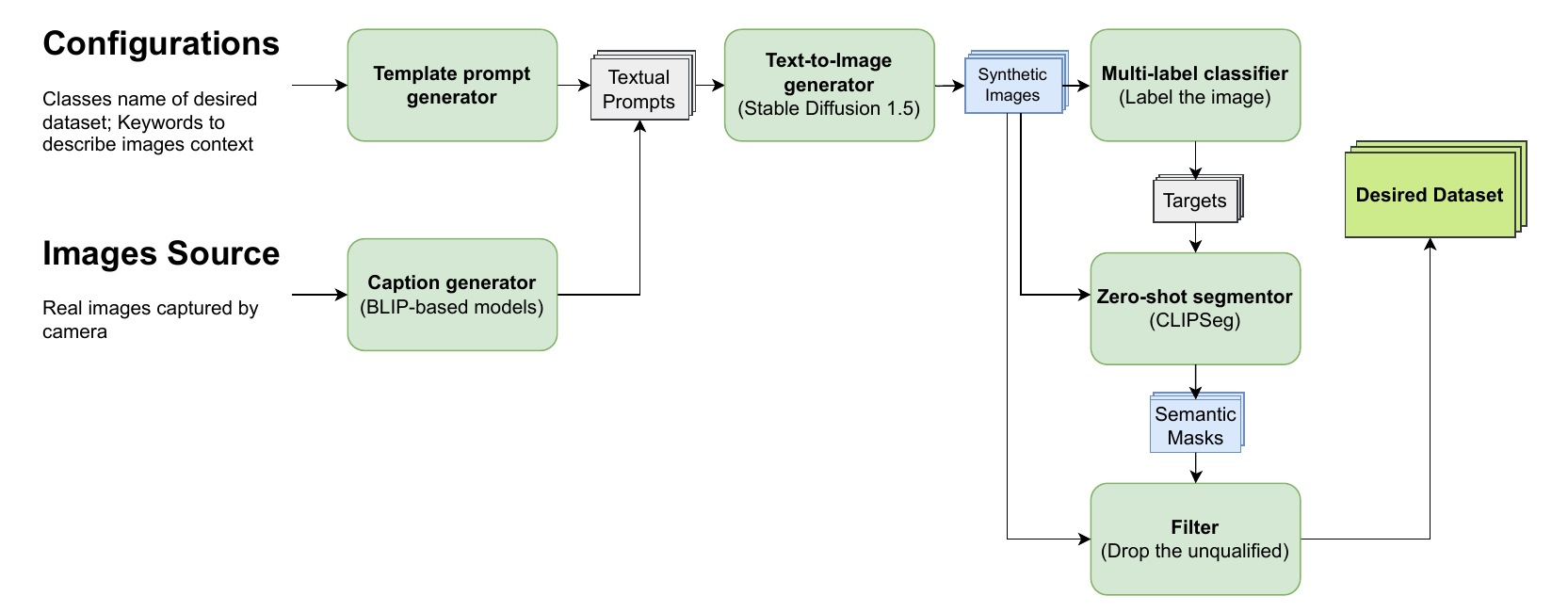}
    \caption{Designed workflow for semantic segmentation dataset synthesis, including three big components: prompting module (leftmost), image generation module (middle), and mask generation module (rightmost).}
    \label{fig:syntagen-overview-pipeline}
\end{figure*}

\subsection{Interactive Flow}

One of the standout features of SynthLab is the ability to create, interact, and customize data processing pipelines. This feature allows users to tailor their workflows to specific requirements and preferences. Users can configure each module independently and adjust inputs and outputs to achieve the desired outcomes. This flexibility ensures that users can design pipelines that are optimized for various tasks, whether they are generating synthetic artistic images, or datasets for the task of extractive like image classification, segmentations, etc. Users can select and sequence different methods, such as Stable Diffusion models, combined with extra models for classification or detection tasks to create complex synthetic images (Figure \ref{fig:pipeline-customization}). The drag-and-drop interface simplifies the process, making it easy for users to arrange and connect different modules, enhancing the overall user experience.

Not only enhance the user experience by providing the drag-and-drop interface, SynthLab provides the advance function to help user export their pipeline with full configuration, store, and reuse later. The exported configuration file can be reloaded back to the interface or in the software devkit (SDK) we provided. For advance users, they can speedup the development progress by just export the optimized pipeline in SynthApp, import to their project and process their data themself privately and actively.

\subsection{System Architecture Implementation}

As SynthLab is a real-time user interactive system, developing a server for many of SynthLab's clients involves creating a robust and efficient system capable of handling and processing data requests to generate predictions using pre-trained machine learning models. This server must ensure low latency and high throughput to handle potentially large volumes of concurrent requests and real-time messages. Key considerations include selecting appropriate hardware and infrastructure, implementing load balancing, ensuring availability and scalability, and integrating with existing data pipelines. Effective server design also involves optimizing model performance through techniques like quantization or model pruning and maintaining a streamlined process for deploying and updating models as they evolve. This section aims to demonstrate in detail how we built the backend of SynthLab and how we solved the problem of a real-time hardware-required system for multiple clients.

SynthLab's server is a single worker whose key components include a Web Server Gateway Interface (WSGI), MongoDB, and Redis. The server is created to execute SynthLab's operations, likely helping clients construct and execute their workflow. Compared to ComfyUI, the closest product's concept to SynthLab, our server is characterized by utilizing multi GPU and serving more than one client with high accessibility and availability. The server of SynthLab is generally a web server, divided into two parts, synchronous and asynchronous. This setup allows for real-time interactions and responses in the app interface via socket connection, while also serving standard HTTP requests for fetching data, and handling authorization actions. 

\section{Usage Scenarios}

SynthLab’s modular design shines in diverse applications, showcasing its ability to tackle real-world challenges across computer vision and beyond. Below, we outline three scenarios that highlight its flexibility, each driven by a specific need to simplify complex tasks, enhance dataset diversity, or extend its utility to new domains.

\subsection{Scenario 1: Synthesizing Regular Object Images with Semantic Masks}
\label{sec:scenario_1}

High-quality annotated datasets are costly and scarce, motivating SynthLab’s pipeline to generate synthetic images with precise semantic masks efficiently (Figure~\ref{fig:syntagen-overview-pipeline}). This multi-stage pipeline integrates prompt preparation, image generation, and mask creation. It begins with a pre-trained CLIP model \cite{radford2021learning} mining datasets like PASCAL VOC to craft detailed captions. These guide Stable Diffusion (SD) \cite{rombach2022highresolution}, a text-to-image model, to produce realistic images aligned with the intended content. To ensure accuracy, a multi-label classifier verifies object presence, filtering out errors, before CLIPSeg \cite{luddecke2022image} generates fine-grained segmentation masks. Post-processing further refines the output, delivering scalable, controlled datasets for training computer vision models efficiently.

\begin{figure*}[t!]
    \centering
    \includegraphics[width=\linewidth]{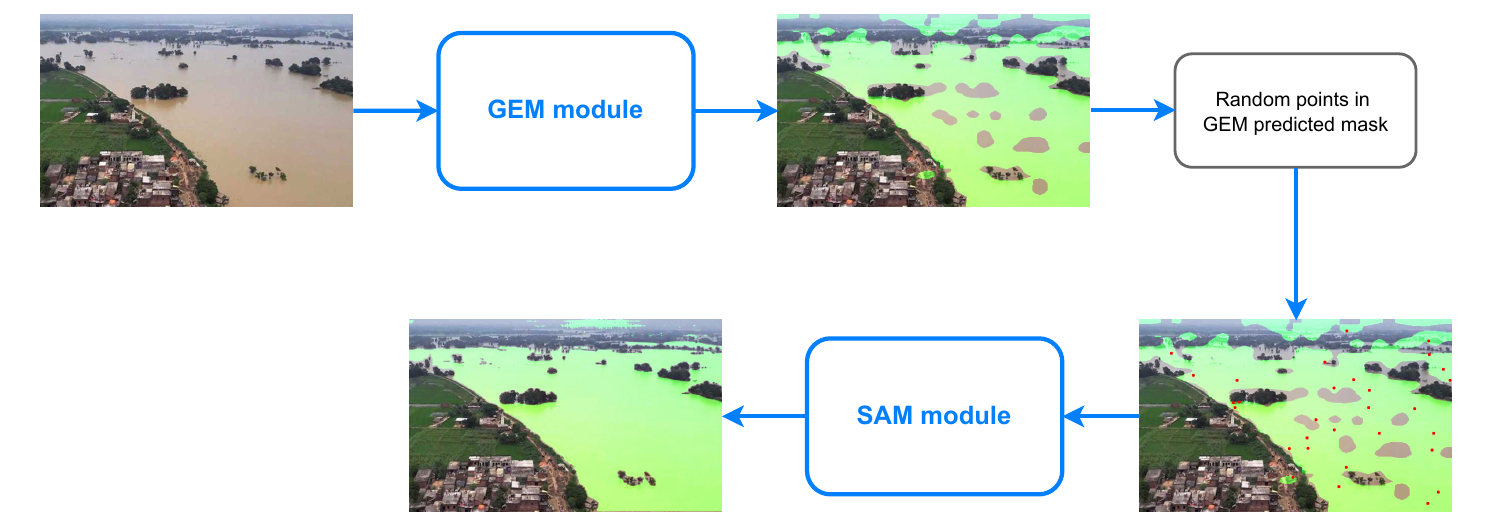}
    \caption{Designed workflow to label uncommon objects.}
    \label{fig:gem_sam}
\end{figure*}

\subsection{Scenario 2: Segmenting Uncommon Objects}
\label{sec:scenario_2}

\begin{figure}[t!]
    \centering
    \subfloat[Flood scenes.\label{fig:flood-samples}]{\includegraphics[width=\linewidth]{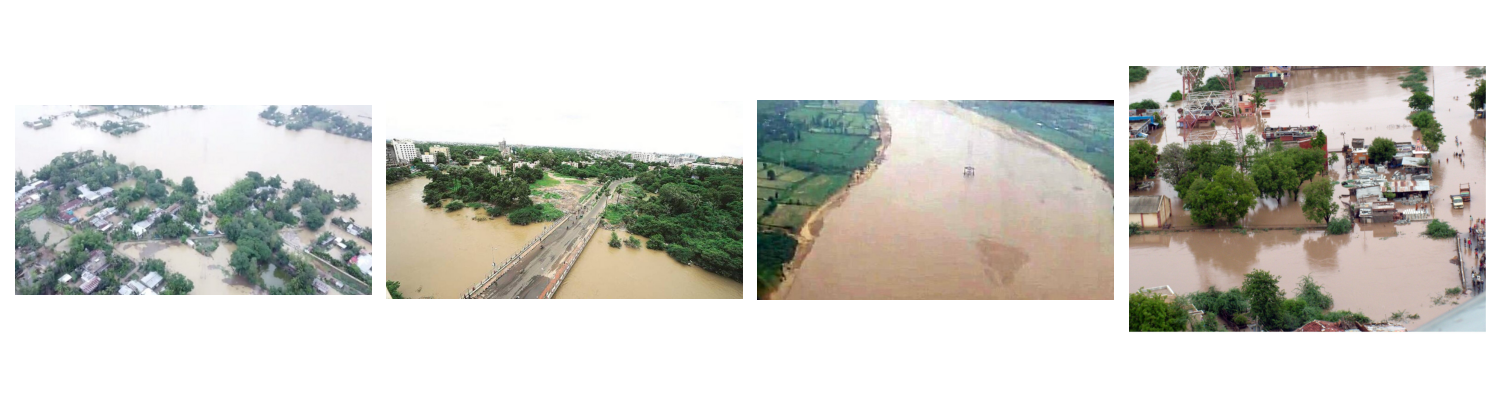}}\\
    \subfloat[Niche contexts.\label{fig:uncommon-objects}]{\includegraphics[width=\linewidth]{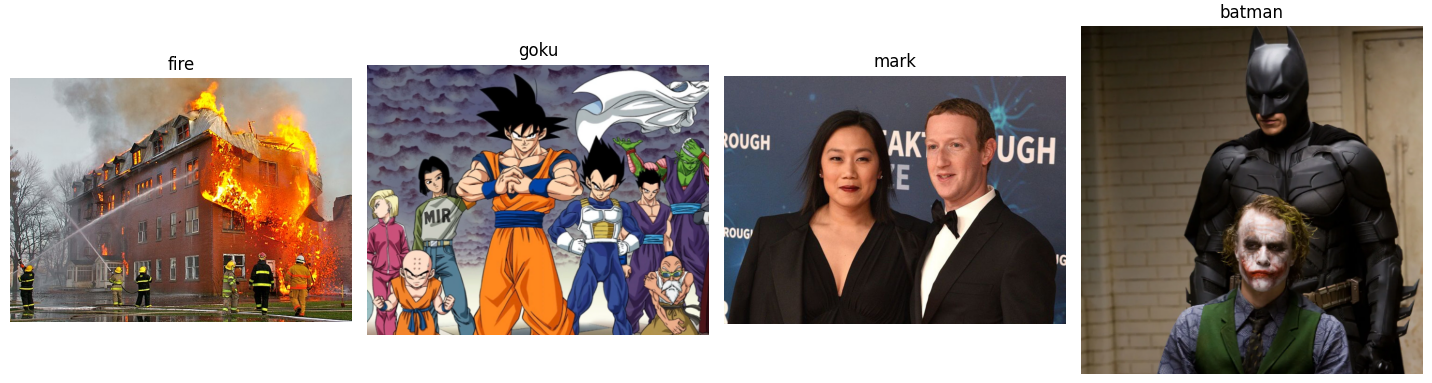}}
    \caption{Examples of uncommon objects.}
    \label{fig:flood-uncommon}
\end{figure}

Traditional models struggle to label rare objects without extensive training data, inspiring SynthLab to enable zero-shot segmentation for uncommon items such as those in flood scenes or niche contexts (Figure \ref{fig:flood-uncommon}). This scenario combines GEM (Grounding Everything Model) \cite{bousselham2023grounding} and SAM (Segment Anything Model) \cite{kirillov2023segment} in a seamless workflow (Figure~\ref{fig:gem_sam}). GEM employs self-attention to cluster and localize objects, think rusted tools or floating debris, outperforming other training-free methods in our tests. SAM then takes over, using $N$ random points from GEM’s initial mask to refine boundaries with precision, even for irregular shapes. This workflow excels in diverse contexts, from disaster imagery to historical scans, ensuring synthetic datasets are both accurate and varied, ideal for training models to handle edge cases effectively.

\subsection{Scenario 3: Visualizing and Interacting with Chemistry}

\begin{figure}[t!]
    \centering
    \includegraphics[width=\linewidth]{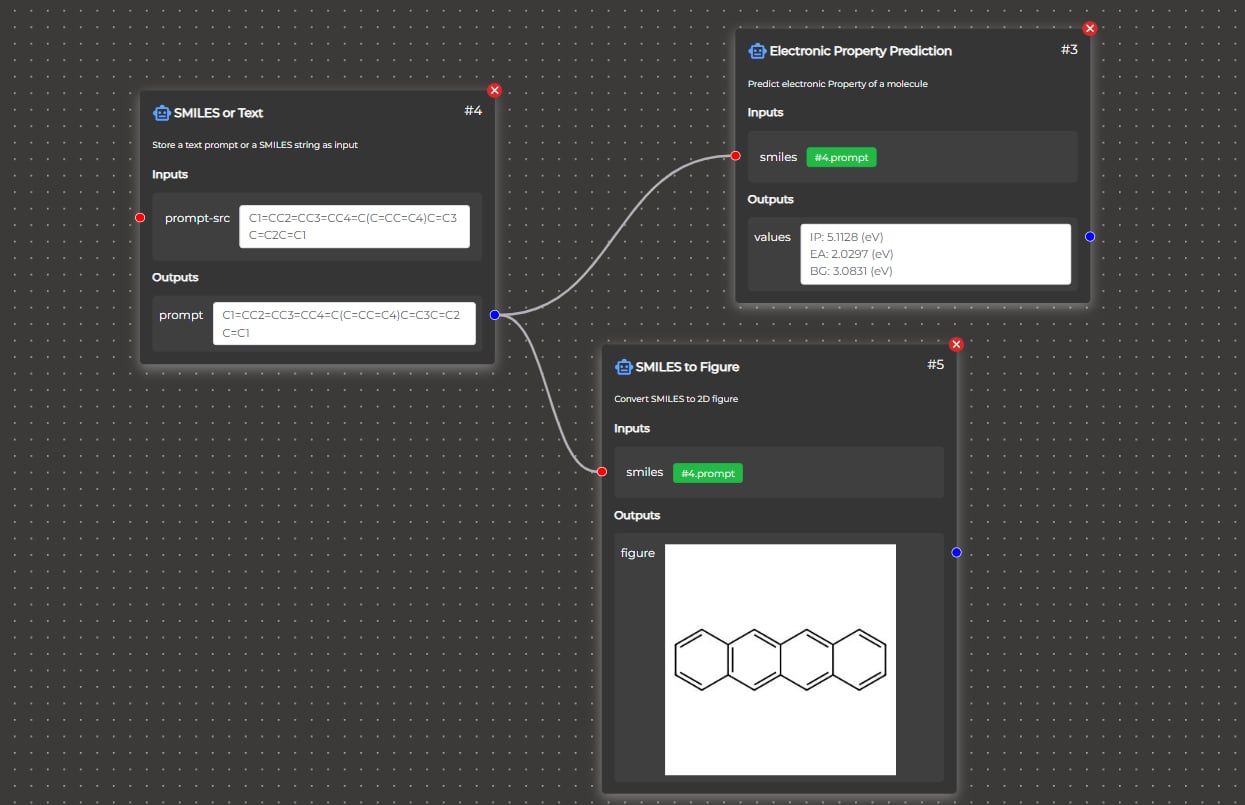}
    \caption{Example of chemistry-related pipeline.}
    \label{fig:chemplay-demo}
\end{figure}

SynthLab’s potential extends beyond pure vision tasks. We built on SynthLab’s core framework with ten specialized modules for chemistry. Figure~\ref{fig:chemplay-demo} demonstrates that SynthLab can support molecule property prediction, visualization, and transformation workflows. With minimal adjustments to the original tech stack, SynthLab proves the platform’s versatility, offering a foundation for chemists and researchers to interact with molecular data, expanding its impact beyond traditional computer vision tasks.

\section{User Study}
\label{sec:user-study}

\begin{table*}[t!]
\caption{Summary of participant questions and responses, reflecting overall impressions. Most participants viewed SynthApp as a promising tool for image processing and generative AI.}
    \label{tab:questionnaire-answer}
\resizebox{\textwidth}{!}{
    \begin{tabular}{clcc} 
    \toprule 
    \textbf{No.} & \textbf{Question} & \textbf{Most response} & \textbf{Other responses} \\
    \midrule
    1 & \makecell[l]{Which features of SynthApp are the most impressive?} & \makecell{The interactive (41\%) } & \makecell{UI-UX (13\%), Integrated Modules (13\%), \\ Workflow Customization (25\%), Other (8\%) } \\ 
    2 & \makecell[l]{Is SynthApp easy to use?} & \makecell{Yes (88\%)} & No (12\%) \\
    3 & \makecell[l]{Who is the most needed SynthApp's app concept?} & \makecell{Professionals in CV (30\%)} & \makecell{Organizations (27\%), Researcher (21\%), \\ Student (13\%),  Other (9\%)} \\ 
    4 & \makecell[l]{Do SynthApp's integrated modules work smoothly together? } & \makecell{Yes (85\%)} & Neutral (12\%), No (3\%) \\ 
    5 & \makecell[l]{Is the instruction and demo easy to follow?} & \makecell{Yes (67\%)} & \makecell{No (9\%), No need (24\%)}\\ 
    6 & \makecell[l]{Is SynthApp good for creativity?} &  \makecell{Yes (73\%)} & \makecell{Maybe (27\%)} \\ 
    7 & \makecell[l]{Do you think SynthApp has the potential to be \\ 
     an effective and powerful tool in generative AI?} & \makecell{Yes (67\%)} & \makecell{Maybe (33\%)} \\ 
    8 & \makecell[l]{Do you think SynthApp is good for daily users?}  & \makecell{Yes (48\%)} & \makecell{Maybe (39\%), No (13\%)} \\ 
    \bottomrule    
    \end{tabular}
}
\end{table*}

To comprehensively evaluate SynthLab’s functionality and practical utility, we conducted a two-part, mixed-methods user study. By engaging a diverse cohort of participants, we aimed to assess the platform's accessibility for novices and its effectiveness for experts.

\subsection{Study \#1: Functionality Evaluation}

The primary objective of our initial study was to assess the overall functionality, general user satisfaction, and intuitiveness of SynthLab’s node-based interface across varying levels of technical expertise

\subsubsection{Participants}

We recruited over 30 participants for this evaluation. The cohort was composed of approximately two-thirds undergraduate computer science students and one-third industry professionals possessing specialized expertise in artificial intelligence, computer vision, and machine learning.

\subsubsection{Metrics}

Participants evaluated the system using a 5-point Likert scale (1: least satisfied, 5: most satisfied) across three core criteria: user-friendly interface (UI-UX), ease of usage (Usability), and application concept. Additionally, we utilized post-study interviews to gather qualitative metrics regarding feature impressions, integration smoothness, and perceived user base.

\subsubsection{Tasks}

Participants were assigned an exploratory task where they were given free rein to navigate the SynthLab interface, experiment with different modules, and familiarize themselves with the tool's drag-and-drop mechanics.

\subsubsection{Apparatus and Procedure}

Participants accessed the web-based SynthLab platform independently. After completing their free exploration of the interface and its available modules, users filled out the quantitative satisfaction survey and participated in brief follow-up interviews to provide deeper contextual insights.

\subsubsection{Results}

\begin{figure}[t!]
  \centering
  \subfloat[Statistical distribution of user satisfaction scores.]{\label{fig:10a}
  \includegraphics[width=\linewidth, trim={0 0 15mm 0}, clip]{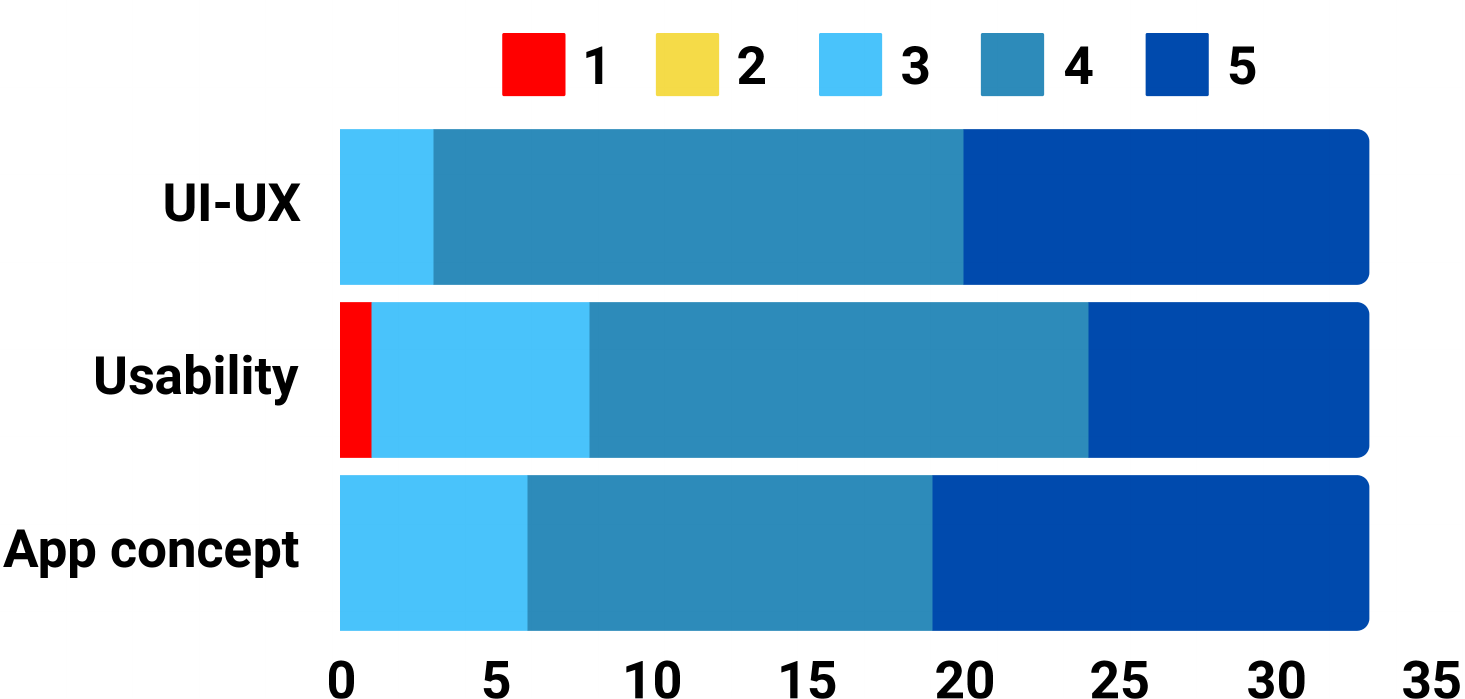}
  }%
  \qquad
  \subfloat[Overall quantitative summary of user satisfaction scores.]{\label{fig:10b}
    \begin{tabular}{lcc}
        \toprule
         \textbf{Criteria} & \textbf{MOS} & \textbf{IQR} \\
        \midrule
        SynthApp's design, UI, and UX & 4.3 & 4 \\ 
        The usability & 4.0 & 4 \\ 
        The app concept & 4.2 & 5 \\ 
        \midrule
        Average score &  4.2 & - \\
        \bottomrule
        \end{tabular}
    }
  \caption{Users’ satisfaction evaluated across three criteria: user-friendly interface, ease of usage, and application concept. Participants were most interested in the UI-UX design and the overall concept of SynthLab. There were almost no scores under three, except for one provided by a user who took the survey without any prior AI experience.}%
\label{fig:SynthLab-user-study}
\end{figure}

The evaluation yielded highly positive feedback, with SynthLab earning a strong overall Mean Opinion Score (MOS) of 4.2 out of 5 (Figure~\ref{fig:SynthLab-user-study}). Specifically, the UI-UX design scored 4.3, the application concept scored 4.2, and general usability scored 4.0. The impressive scores reflect highly positive overall user experiences with our platform.

Survey results indicated broad acceptance: 88\% of users found the application easy to use, and 85\% reported that the integrated modules worked smoothly together (Table~\ref{tab:questionnaire-answer}). Qualitative interviews revealed a slight learning curve for beginners regarding technical module nomenclature, while expert users expressed a desire for enhanced module configurability. Notably, 80\% of respondents identified researchers, organizations, and computer vision professionals as the primary beneficiaries of the platform, highlighting its broad potential for general and professional use.




\subsection{Study \#2: Practical Usage Evaluation}
\label{sec:pilot-study}


Building on the first study, this evaluation aimed to assess SynthLab’s effectiveness in solving complex, real-world problems. We specifically focused on evaluating its ability to simplify intricate data-annotation and generative AI workflows compared to existing baseline tools and manual methods.

\subsubsection{Participants}

We utilized a purposive subset of seven participants, all of whom possessed prior familiarity with ComfyUI, a popular baseline node-based tool. This group included three undergraduate students with no prior AI experience (novices) and four graduate students or industry professionals with applied AI experience (experts).

\subsubsection{Tasks}

Participants engaged in two distinct experimental tasks:

\begin{itemize}
    \item \textbf{Task 1 (Image Generation):} Participants were asked to generate images matching specific textual descriptions using both the baseline tool (ComfyUI) and SynthLab.
    \item \textbf{Task 2 (Dataset Synthesis):} Participants were challenged to construct a 1,000-sample semantic segmentation dataset spanning four object categories (cars, bicycles, motorbikes, and trucks). They first attempted this manually using existing online resources and tools, and then repeated the task using SynthLab’s integrated pipeline.
\end{itemize}

\subsubsection{Apparatus and Procedure}

Participants engaged in supervised, six-hour evaluation sessions. To ensure consistent processing performance and eliminate hardware-based confounding variables, uniform hardware was provided to all users. The procedure began with a 10–15 minute setup period for ComfyUI, followed by a 20–30 minute warm-up phase allowing users to familiarize themselves with both interfaces. During the experimental tasks, a host closely monitored the participants, providing necessary technical support and conceptual guidance, which proved especially beneficial for the novice undergraduates navigating complex AI pipelines.

\subsubsection{Results}

Observations from Task 1 revealed that while expert users could seamlessly leverage their ComfyUI experience to craft detailed outputs, novices struggled significantly with model setup and parameter tuning in the baseline tool. Task 2 highlighted severe bottlenecks in manual dataset creation; participants discovered that online datasets often lacked the required classes or were provided in incompatible formats, stalling their progress entirely. Conversely, SynthLab’s pipeline architecture streamlined the process significantly. By utilizing zero-shot models like Grounded-SAM, participants dynamically generated precise masks directly from raw images, vastly outperforming their manual efforts. Despite novices noting a learning curve related to limited documentation, the system dramatically empowered non-experts. Most notably, one undergraduate student with minimal AI experience successfully combined multiple advanced modules (GroundingDINO, OWLv2, and SAM) to generate a high-confidence, 1,000-sample dataset in approximately five hours-an exceptional achievement for his skill level. Ultimately, participants lauded SynthLab for not only boosting raw productivity but also effectively organizing and managing complex data workflows.

\section{Experiments}

\subsection{Synthesizing Regular Object Images with Semantic Masks}

\begin{figure}[t!]
    \centering
    \includegraphics[trim={25mm 140mm 0 0}, clip, width=\linewidth]{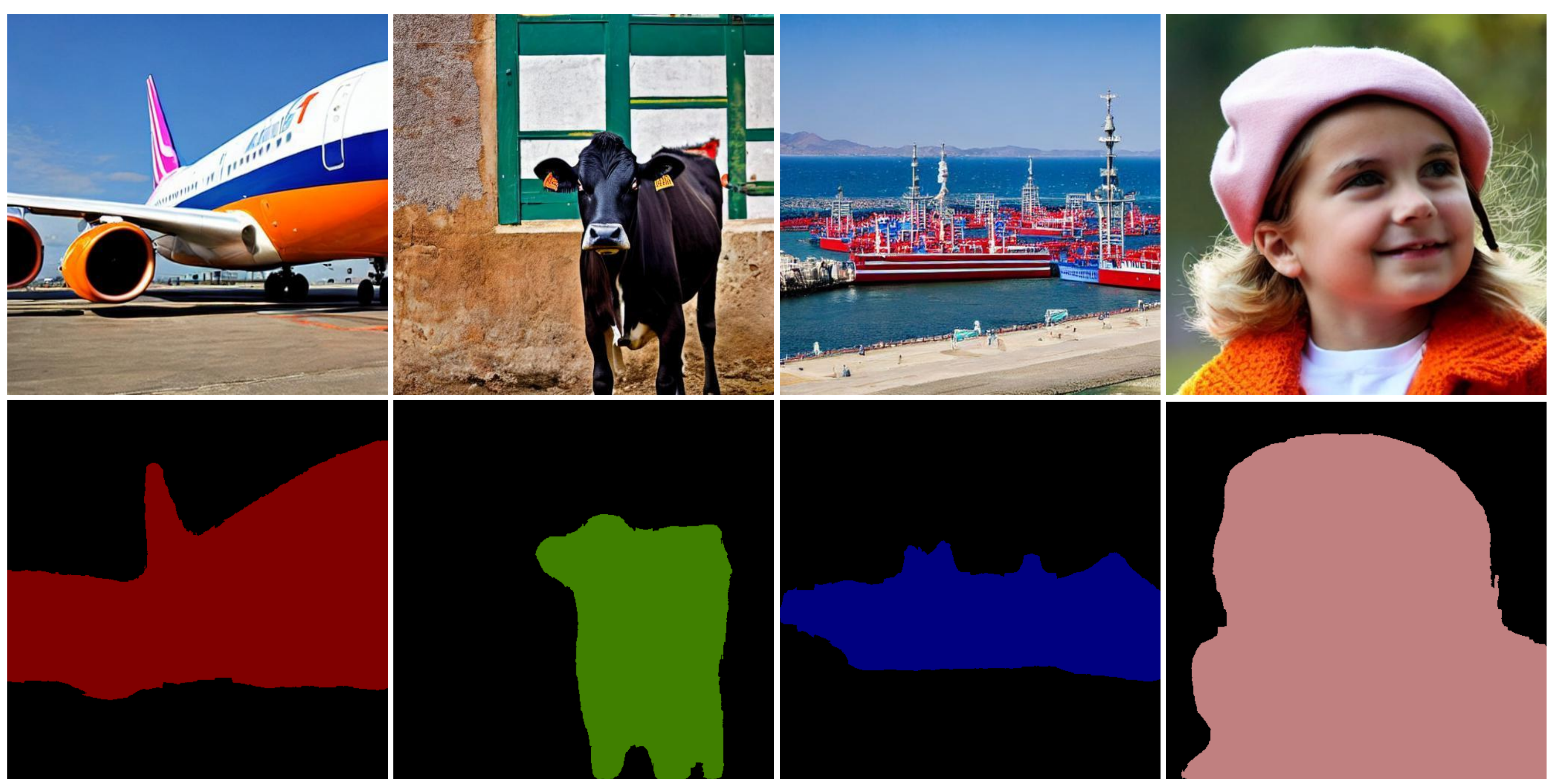}
    \hfill
    \includegraphics[trim={25mm 140mm 0 0}, clip, width=\linewidth]{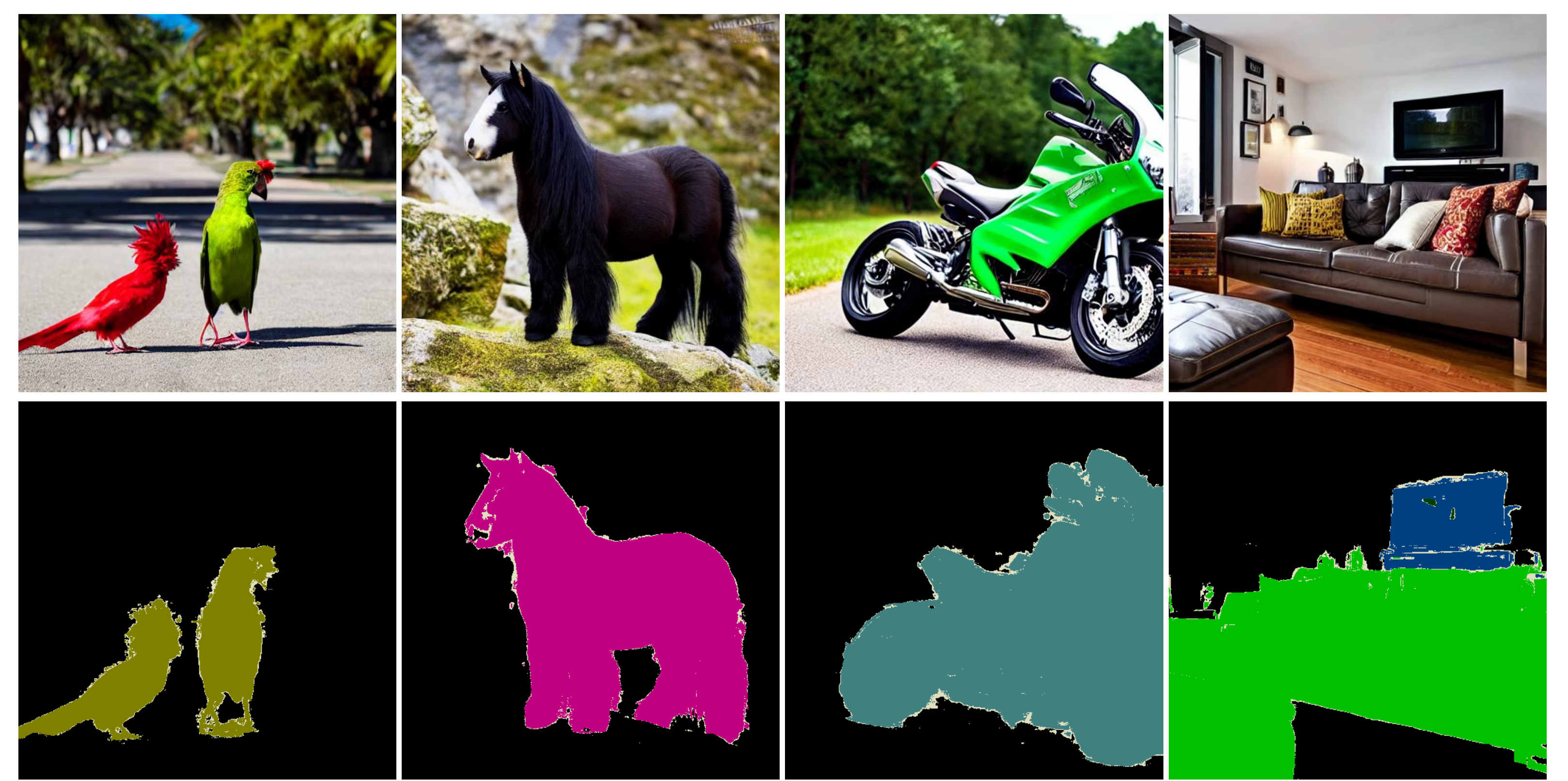}
    \caption{Generated dataset, including images and corresponding semantic masks.}
    \label{fig:mask-generation-compare}
\end{figure}

\begin{table}[t!]
\centering
\caption{Evaluation results of usage of synthesis dataset for training semantic segmentation models (mIoU: Higher is better).}
\begin{tabular}{lc} 
\toprule
\textbf{Method} & \textbf{mIoU (\%)} \\
\midrule

Stillger et al.~\cite{Stillger_2024_CVPRW} & 46.25 \\  
Nguyen et al.~\cite{nguyen2023dataset} & 46.85 \\ 
Wang et al.~\cite{Wang_2024_CVPRW} & 47.36 \\ 
\textbf{Ours} & \textbf{51.61} \\ 
\bottomrule
\end{tabular}

\label{tab:exp-compare}
\end{table}

To evaluate the creation of high-quality synthetic datasets, we leverage only 20 class names from the PASCAL VOC~\cite{everingham2015pascal} for train semantic segmentation models. Using our proposed pipeline in Section~\ref{sec:scenario_1}, we generated a synthetic dataset, which includes 8049 image/mask pairs. Our generated dataset (illustrated in Fig.~\ref{fig:mask-generation-compare}) and other synthesis datasets~\cite{nguyen2023dataset, Stillger_2024_CVPRW, Wang_2024_CVPRW} were used to train the same DeepLabV3 model~\cite{chen2017rethinking} with Resnet-50~\cite{He_2016_CVPR} backbone, using a batch size of 8 for a maximum of 10000 iterations, and subsequently assessing its performance on the validation set of PASCAL VOC. The results in Table~\ref{tab:exp-compare} show that our approach significantly outperforms existing solutions.

\subsection{Segmenting Uncommon Objects}

\begin{table}[t!]
\centering
\caption{Results of flood segmentation on the flood dataset with different methods.}
\label{tab:uncommon-result} 
\begin{tabular}{lc} 
\toprule
\textbf{Method} & \textbf{mIoU (\%)} \\
\midrule 
DINO~\cite{liu2024grounding} + SAM~\cite{kirillov2023segment} & 3.32 \\ 
CLIPSeg~\cite{luddecke2022image} & 21.25 \\ 
GEM~\cite{bousselham2023grounding} & 53.25 \\
\textbf{GEM + SAM (Ours)} & \textbf{73.17} \\ 
\bottomrule
\end{tabular}
\end{table}

\begin{figure*}[t!]
    \centering
    \includegraphics[width=\textwidth]{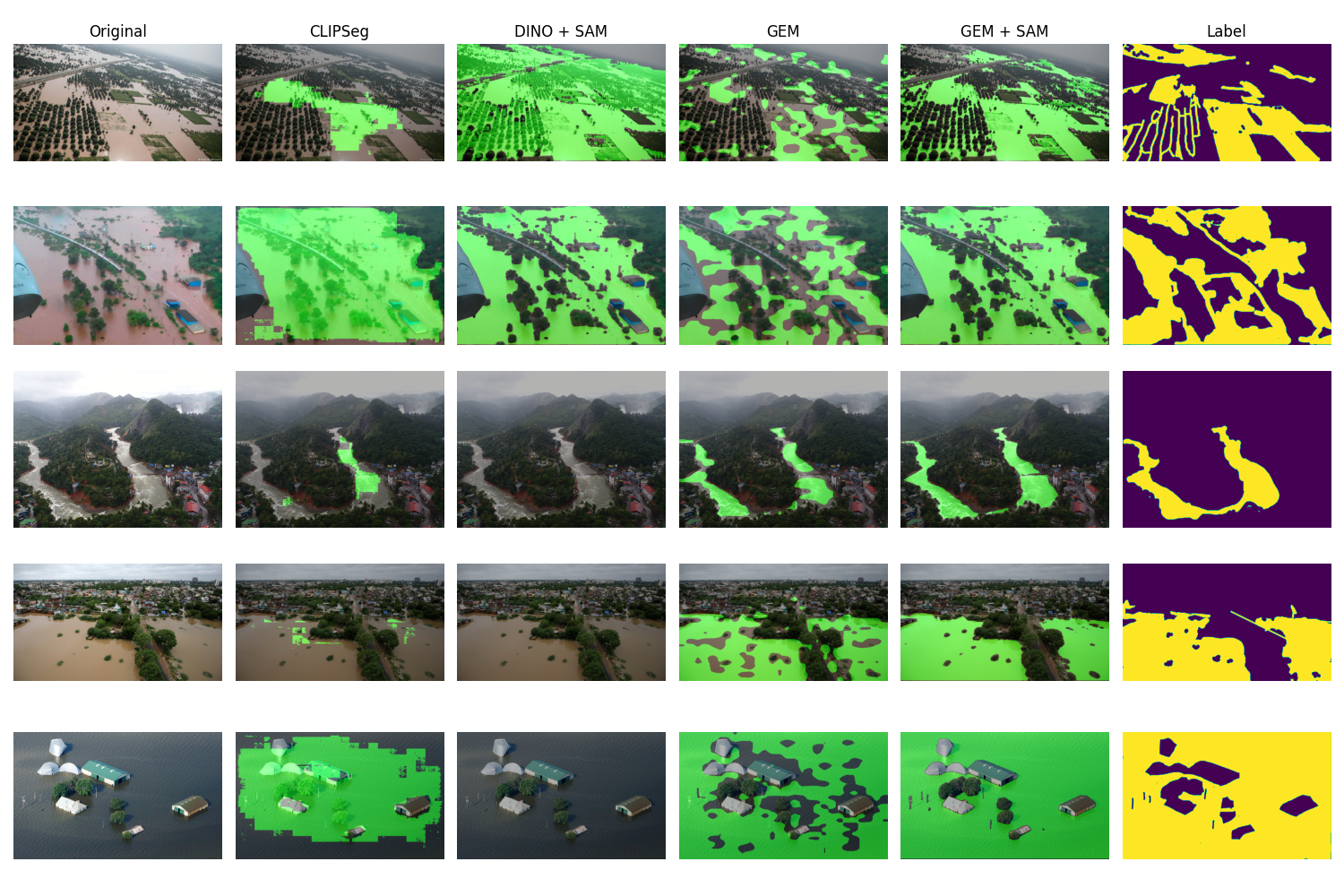}
    \caption{Qualitative results of flood segmentation across different open-vocabulary methods. Our proposed approach (GEM+SAM) most closely matches the ground truth labels. While methods like DINO + SAM and CLIPSeg struggle to accurately segment these uncommon objects, utilizing SAM as a boosting module to refine GEM's initial predictions yields highly accurate masks.}
    \label{fig:flood-results}
\end{figure*}

To evaluate the effectiveness of our approach in segmenting uncommon objects in Section~\ref{sec:scenario_2}, we compared our approach against other popular open-vocabulary segmentation methods in the flood segmentation dataset\footnote{\url{https://www.kaggle.com/datasets/lihuayang111265/flood-semantic-segmentation-dataset}}. We also compared with GEM \cite{bousselham2023grounding} only to validate the effects of SAM \cite{kirillov2023segment}. As shown in Table~\ref{tab:uncommon-result}, our approach significantly outperforms other methods. SAM is a widely used method by many developers since it's lightweight and robust, but as a result, the method failed significantly in labeling segmentation masks for uncommon objects. CLIPSeg \cite{luddecke2022image} shows it is better than SAM in this task but still far away from our proposed method. GEM only can perform a good result but with the SAM module working as a boost module, the mIoU score is gratefully increased.



\section{Conclusion and Future Work}

In this paper, we present a robust and adaptable platform for addressing complex computer vision challenges, particularly in synthetic dataset generation for semantic segmentation. Modular architecture of our SynthLab platform empowers users to craft tailored pipelines for diverse tasks. The ability to independently modify each module and debug complex pipelines ensures that users can optimize their workflows precisely and efficiently. Delivered through an intuitive interface, SynthLab democratizes access to advanced data synthesis tools, appealing to users at all levels of expertise. 

Future work will focus on improving SynthLab’s usability and versatility. Planned enhancements include a more streamlined user interface to reduce cognitive load and expanded support for diverse data modalities (e.g., audio for speech tasks, video for summarization). 

\bibliographystyle{ACM-Reference-Format}
\balance
\bibliography{reference}

\end{document}